\documentclass{article}

\usepackage{microtype}
\usepackage{booktabs} 

\usepackage{hyperref}

\usepackage[accepted]{icml2023}

\usepackage{amssymb}
\usepackage{mathtools}
\usepackage{amsthm}
\usepackage{bm}

\usepackage[capitalize,noabbrev]{cleveref}

\usepackage{makecell}
\newcommand{\ra}[1]{\renewcommand{\arraystretch}{#1}}

\usepackage{dsfont}

\usepackage[pdftex]{graphicx}

\usepackage{enumitem}
\usepackage{wrapfig}
\usepackage{gensymb}
\usepackage{xparse} 
\usepackage{caption}
\usepackage{multicol}
\usepackage{subcaption}
\usepackage{amsmath}
\usepackage[algo2e,ruled,vlined]{algorithm2e}
\SetKw{Break}{break}

\usepackage{hyperref}

\newcommand{\ie}{\textit{i}.\textit{e}., }
\newcommand{\eg}{\textit{e}.\textit{g}. }

\newcommand{\changed}[1]{}

\theoremstyle{plain}

\theoremstyle{definition}

\theoremstyle{remark}

\usepackage[textsize=tiny]{todonotes}

\icmltitlerunning{VectorMapNet: End-to-end Vectorized HD Map Learning}

\begin{document}

\twocolumn[
\icmltitle{VectorMapNet: End-to-end Vectorized HD Map Learning}




\begin{icmlauthorlist}
\icmlauthor{Yicheng Liu}{yyy,tsinghua}
\icmlauthor{Tianyuan Yuan}{tsinghua}
\icmlauthor{Yue Wang}{mit}
\icmlauthor{Yilun Wang}{liauto}
\icmlauthor{Hang Zhao}{tsinghua,yyy} \\
\url{https://tsinghua-mars-lab.github.io/vectormapnet/}

\end{icmlauthorlist}

\icmlaffiliation{yyy}{Shanghai Qi Zhi Institute}
\icmlaffiliation{tsinghua}{Tsinghua University}
\icmlaffiliation{mit}{MIT}
\icmlaffiliation{liauto}{Li Auto}

\icmlcorrespondingauthor{Hang Zhao}{ZhaoHang0124@gmail.com}

\icmlkeywords{Map Learning, Autonomous Driving, Vectorization}

\vskip 0.3in
]



\printAffiliationsAndNotice{}  

\begin{abstract}
Autonomous driving systems require High-Definition (HD) semantic maps to navigate around urban roads. Existing solutions approach the semantic mapping problem by offline manual annotation, which suffers from serious scalability issues.  Recent learning-based methods produce dense rasterized segmentation predictions to construct maps. However, these predictions do not include instance information of individual map elements and require heuristic post-processing to obtain vectorized maps. To tackle these challenges, we introduce an end-to-end vectorized HD map learning pipeline, termed VectorMapNet. VectorMapNet takes onboard sensor observations and predicts a sparse set of polylines in the bird's-eye view. This pipeline can explicitly model the spatial relation between map elements and generate vectorized maps that are friendly to downstream autonomous driving tasks. Extensive experiments show that VectorMapNet achieve strong map learning performance on both nuScenes and Argoverse2 dataset, surpassing previous state-of-the-art methods by 14.2 mAP and 14.6mAP. Qualitatively, VectorMapNet is capable of generating comprehensive maps and capturing fine-grained details of road geometry. 
To the best of our knowledge, VectorMapNet is the first work designed towards end-to-end vectorized map learning from onboard observations. 
\end{abstract}

\vspace{-2em}
\section{Introduction}
Autonomous driving systems require an understanding of map elements on the road, including lanes, pedestrian crossing, and traffic signs, to navigate around the world. Such map elements are typically provided by pre-annotated High-Definition (HD) semantic maps in existing pipelines~\citep{rong2020lgsvl}. However, these methods face scalability issues due to their heavy reliance on human labor for annotating HD maps. Additionally, they necessitate precise localization of the ego-vehicle to derive local maps from the global one, a process that could introduce meter-level errors.

In contrast, our focus lies in developing a learning-based approach for online HD semantic map learning. The aim is to use onboard sensors, including LiDARs and cameras, to estimate map elements on-the-fly. This methodology avoids the need for localization, allowing for prompt updates. Furthermore, learning-based methods can generate uncertainty or confidence indicators that downstream modules, such as motion forecasting and planning, can utilize to offset imperfect perception. These methods can leverage increasing data and model size, promptly reflect current conditions, and generalize from annotated maps to under-annotated or even non-annotated areas (please refer to Figure~\ref{fig:qualitative_example}).

Most of HD semantic map learning methods~\citep{li2021hdmapnet,philion2020lift,roddick2020predicting,zhou2022cross} consider the task as a semantic segmentation problem in bird's-eye view (BEV), which rasterizes map elements into pixels and assigns each pixel with a class label. This formulation makes it straightforward to leverage fully convolutional networks.
However, rasterized maps are not an ideal map representation for autonomous driving, for three reasons.
First, rasterized maps lack instance information necessary to distinguish map elements with the same class label but different semantics, \eg left boundary and right boundary.
Second, it is hard to enforce spatial consistency within the predicted rasterized maps, \eg nearby pixels might have contradicted semantics or geometries.
Third, 2D rasterized maps are incompatible with most autonomous driving systems which consume instance-level 2D/3D vectorized maps for motion forecasting and planning. 

To alleviate these issues and produce vectorized outputs, HDMapNet~\citep{li2021hdmapnet} generates semantic, instance, and directional maps and vectorizes these three maps with a hand-designed post-processing algorithm. However, HDMapNet still relies on the rasterized map predictions, and its heuristic post-processing step  restricts the model's scalability and performance. 

\begin{figure*}[tp]
    \centering
    \includegraphics[width=\linewidth]{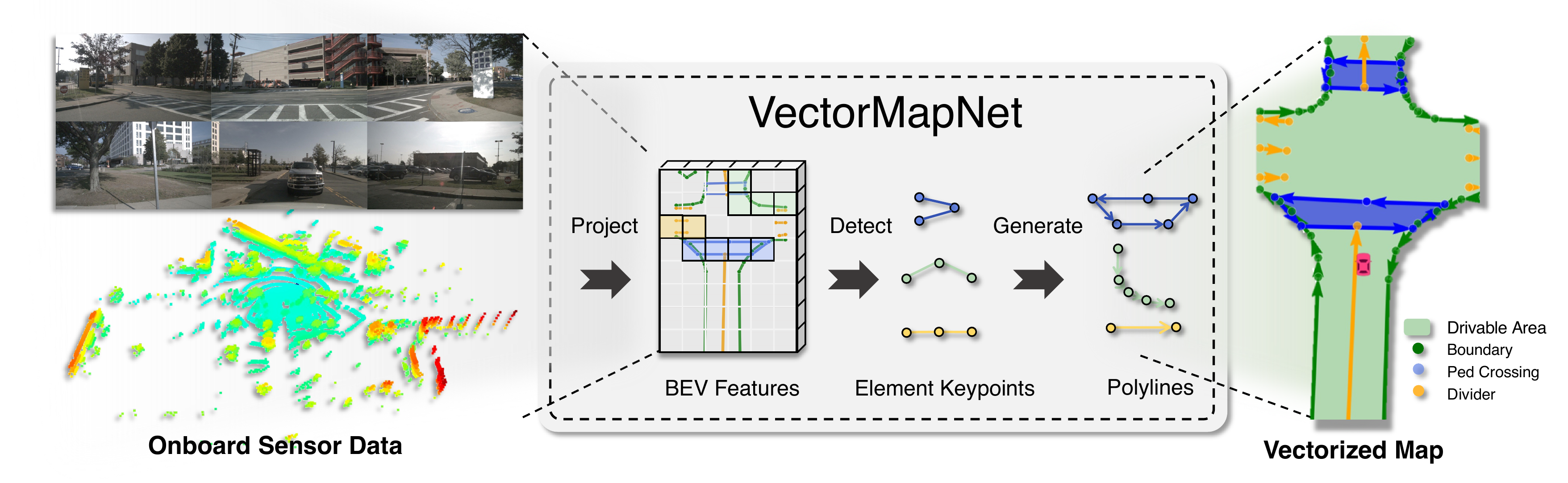}
    \vspace{-1em}
    \caption{
      An overview of VectorMapNet. Sensor data is encoded to BEV features in the same coordinate as map elements. VectorMapNet detects the locations of map elements from BEV features by leveraging element queries. The vectorized HD map is built upon a sparse set of polylines that are generated from the detection results. Since our polylines are directional, we can infer drivable area of a map.
    }
    \vspace{-1.5em}
    \label{fig:teaser}
\end{figure*}

In this paper, we propose an end-to-end vectorized HD map learning model named VectorMapNet, an end-to-end framework that does not involve dense semantic pixels or sophisticated post-processing steps. Instead, it represents map elements as a set of polylines closely related to downstream tasks, \eg motion forecasting~\citep{vectornet}. Therefore, the mapping problem boils down to predicting a sparse set of polylines from sensor observations. Specifically, we pose it as a detection problem and leverage recent set detection and sequence generation methods. First, VectorMapNet aggregates features generated from different modalities (\eg camera images and LiDAR) into a common BEV feature space. Then, it detects map element locations based on learnable element queries and BEV features. Finally, we decode each element query into a polyline. An overview of VectorMapNet is shown in Figure~\ref{fig:teaser}.

Our experiments show that VectorMapNet achieves state-of-the-art performance on the public nuScenes dataset~\citep{caesar2020nuscenes} and Argoverse2~\citep{Argoverse2}, outperforming HDMapNet and another baseline by at least 14.2 mAP. 
Qualitatively, VectorMapNet builds a more comprehensive map than previous works and can capture fine details, \eg jagged boundaries.
Furthermore, we feed our predicted vectorized HD map into a downstream motion forecasting module, demonstrating the predicted map's compatibility and effectiveness. To summarize, the contributions of the paper are as follows:
\begin{itemize}[leftmargin=*]
    \vspace{-1em}
    \setlength\itemsep{0em}
    \item We present VectorMapNet, an end-to-end mapping approach that eliminates the need for map rasterization and post-processing by predicting vectorized outputs directly from sensor observations.
    \item We utilize polyline, a flexible primitive with variable lengths and encoded order, to accommodate the heterogeneous nature of map elements. This approach effectively formulates the construction of a polyline map as a detection issue, thereby introducing a new strategy to the mapping paradigm.
    \item We adapt detection transformer (DETR) models to locate deformable elements within a 3D space. Recognizing that prevalent centerpoint-based feature extraction methods fall short when dealing with map elements of varying sizes and shapes, we propose an innovative solution. Our novel method overcomes these limitations, delivering state-of-the-art performance in online semantic HD map learning tasks.
\end{itemize} 

\begin{figure*}[htp]
    \centering
    \includegraphics[width=\linewidth]{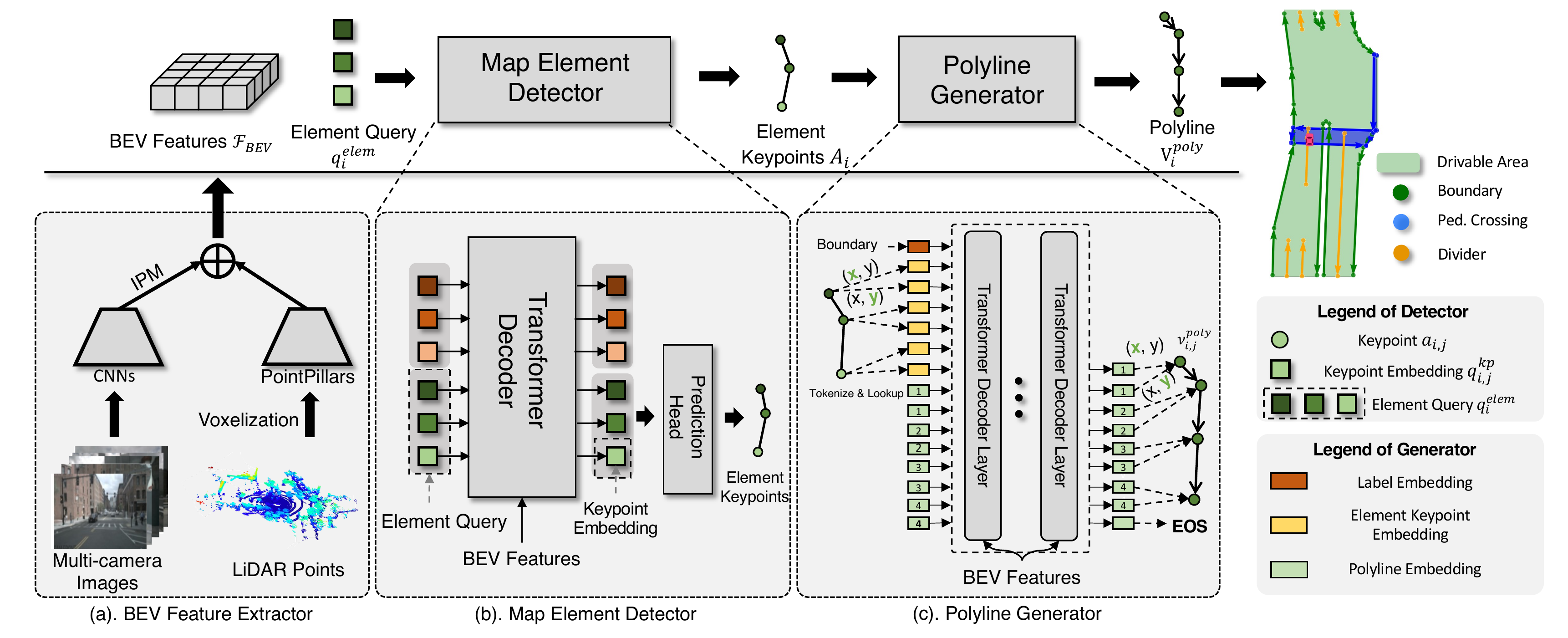}
    \caption{The network architecture of VectorMapNet. The top row is the pipeline of VectorMapNet generating polylines from raw sensor inputs. The bottom row illustrates detailed structures and inference procedures of three primary components of VectorMapNet: BEV feature extractor, map element detector, and polyline generator. Numbers in polyline embeddings indicate predicted vertex indexes.
    }
    \label{fig:overview}
    \vspace{-1em}
\end{figure*}

\section{Related Works}
\noindent\textbf{Semantic map learning.}
Annotating semantic maps attracts plenty of interests thanks to autonomous driving.  Recently, semantic map learning is formulated as a semantic segmentation problem~\citep{mattyus2015enhancing} and is solved by using aerial images~\citep{mattyus2016hd}, LiDAR points~\citep{yang2018hdnet}, and HD panorama~\citep{wang2016torontocity}. The crowdsourcing tags~\citep{wang2015holistic} are used to improve the performance of fine-grained segmentation.
Instead of using offline data, recent works focus on understanding BEV semantics from onboard camera images~\citep{lu2019monocular,yang2021projecting}, and videos~\citep{can2020understanding}.
Only using onboard sensors as model input is particularly challenging as the inputs and target map lie in different coordinate systems. Recently, several cross-view learning approaches~\citep{philion2020lift,pan2020cross,li2021hdmapnet,zhou2022cross,wang2022detr3d,chen2022futr3d} leverage the geometric structure of scenes to mitigate the mismatch between sensor inputs and BEV representations. 
Some methods~\citep{casas2021mp3, sadat2020perceive} use pixel-level semantic maps to solve downstream tasks, but the entire downstream pipeline needs to be redesigned to accommodate these rasterized map inputs.
Beyond pixel-level semantic maps, our work extracts a consistent vectorized map around ego-vehicle from surrounding cameras or LiDARs, which suits for existing downstream tasks like motion forecasting~\citep{vectornet,tnt,liu2021multimodal} without further post-processing.

\noindent\textbf{Lane detection.}
Lane detection aims to separate lane segments from road scenes precisely. Most lane detection algorithms~\citep{pan2018spatial,neven2018towards} use a pixel-level segmentation technique combined with sophisticated post-processing. Another line of work leverages the predefined proposal to achieve high accuracy and fast inference speed. These methods typically involve handcrafted elements such as vanishing points~\citep{lee2017vpgnet}, polynomial curves~\citep{van2019end}, line segments~\citep{li2019line}, and B\'ezier curves~\citep{feng2022rethinking} to model proposals. In addition to using perspective view cameras as inputs, \citep{homayounfar2018hierarchical} and \citep{liang2019convolutional} extract lane segments from overhead highway cameras and LiDAR imagery with a recurrent neural network.
Instead of discovering the road's topology via boundaries detection, STSU~\citep{can2021structured} and LaneGraphNet~\citep{zurn2021lane} construct lane graphs from centerline segments that are encoded by B\'ezier curves and line segments, respectively.
To model complex geometries in the urban environment, we leverage polylines to represent all the map elements in perceptual scopes. 

\noindent\textbf{Geometric data modeling.}
Another line of work closely related to VectorMapNet is geometric data generation. These methods typically treat geometric elements as a sequence, such as primitive parts of furniture~\citep{li2017grass,mo2019structurenet}, states of sketch strokes~\citep{ha2017neural}, vertices of $n$-gon mesh~\citep{nash2020polygen}, and parameters of SVG primitives \citep{carlier2020deepsvg}.
These methods generate these sequences by leveraging autoregressive models~(\eg Transformer). 
Since the directly modeling sequence is challenging for long-range centerline maps, HDMapGen~\citep{mi2021hdmapgen} views the map as a two-level hierarchy. 
It produces a global and local graph separately with a hierarchical graph RNN. Instead of treating geometric elements as a sequence generation problem, LETR~\citep{xu2021line} models line segment as a detection problem and tackle it with a query-based detector. 
Unlike the above approaches that focus on single-level geometric modelings, such as scene level (\eg line segments in an image) or object-level (\eg furniture), VectorMapNet is designed to address both the scene level and object level geometric modeling. Specifically, VectorMapNet constructs a map by modeling the global relationship between map elements in the scene and the local geometric details inside each element.

\noindent\textbf{Learning vector representations from images.}
VectorMapNet bears some similarities with predicting vector graphics from raster images. Several recent works~\citep{carlier2020deepsvg,reddy2021im2vec} use different vector representations to generate vector images. \citep{ganin2021computer} converts images to CAD, CanvasVAE~\citep{yamaguchi2021canvasvae} learns vectorized canvas layouts from images, and \citep{liu2022end} generates vectorized stroke primitives from a raster line drawing.
The instance segmentation community has also been concerned with a similar task of detecting object contours in a vector form from an image. These methods~\citep{acuna2018efficient,liang2020polytransform,castrejon2017annotating,zorzi2022polyworld,zhang2019jointnet} initialize a contour for every object instance and then refine the vertex positions of the contour.
However, The above methods are highly domain-dependent, and it is non-trivial to adapt them for our task that requires detecting and generating map elements with different semantics and geometry in the 3D world.


\section{VectorMapNet}
\label{sec:VectorMapNet}
\noindent\textbf{Problem Formulation and Challenges.}
Similar to HDMapNet~\citep{li2021hdmapnet}, our task is to vectorize map elements using data from onboard sensors of autonomous vehicle, such as RGB cameras and/or LiDARs. These map elements include but are not limited to: \textit{Road boundaries} (boundaries of roads separating roads and sidewalks, typically irregularly-shaped curves of arbitrary lengths), \textit{Lane dividers} (boundaries dividing lanes on the road, usually straight lines), and \textit{Pedestrian crossings} (regions with white markings indicating legal pedestrian crossing points, typically represented as polygons). 
While the task is clearly defined, it is fraught with complexities and unique challenges when tackling it.
(1) The diverse geometric structures of map elements make it difficult to establish a unified geometric representation. 
(2) The inputs and outputs of the mapping problem are not perfectly aligned. They exist in different view spaces (\eg camera data is in perspective view and map elements are in BEV), and not all map elements are fully visible from input sensors. In some extreme cases, map elements may be completely occluded by vehicles. 
(3) The task requires more than simple vectorization; it also necessitates scene understanding because of the complex geometrical and topological relationships between map elements. For instance, map elements may overlap, or two traffic cones connected with a wire might indicate a road boundary.

\subsection{Method Overview} 
\label{sec:model_overview}
The challenges above underline the need for a primitive that  effectively represents a variety of geometric structures and a model that is capable of capturing the geometrical and topological relationships from various sensor inputs.

\noindent\textbf{Polyline representation.}
The heterogeneous geometry of map elements calls for a unified vectorized representation. We opt to use $N$ polylines $\bm{\mathcal{V}}^{\mathrm{poly}}=\{\bm{V}_1^{\mathrm{poly}},\dots, \bm{V}_N^{\mathrm{poly}}\}$ as primitives to represent these map elements in a map $\mathcal{M}$. Each polyline $\bm{V}^{\mathrm{poly}}_i = \{\bm{v}_{i,n}\in\mathbb{R}^2| n=1,\dots,N_v\}$ is a collection of $N_v$ ordered vertices $\bm{v}_{i,n}$. In practice, we pre-process public autonomous driving semantic maps to obtain a unified polyline representation of map elements: polygons are represented as closed polylines; curves are converted into polylines by applying the Ramer–Douglas–Peucker algorithm~\citep{Ramer1972AnIP}.

Using polylines to represent map elements has three main advantages: 
(1) HD maps are typically composed of a mixture of different geometries, such as points, lines, curves, and polygons. Polylines are a flexible primitive that can represent these geometric elements effectively. 
(2) The order of polyline vertices is a natural way to encode the direction of map elements, which is vital to driving. 
(3) The polyline representation has been widely used by downstream autonomous driving modules, such as motion forecasting~\citep{vectornet}.

\noindent\textbf{VectorMapNet.} We introduce VectorMapNet, an end-to-end model designed to represent a map $\mathcal{M}$ with a sparse set of polylines $\bm{\mathcal{V}}^{\mathrm{poly}}$, thus formulating the task as a sparse set detection problem. In our approach, we convert sensor data into a canonical Bird's Eye View (BEV) representation, $\bm{\mathcal{F}}_{\mathrm{BEV}}$, and model polylines based on this BEV. Given the complexity and diversity of map elements' structural and location patterns and relationships, we divide the task into three distinct components: (1) A \textit{BEV feature extractor} (\S~\ref{sec:bev_backbone}) that lifts various sensor modality inputs into a canonical feature space.
(2) A \textit{map element detector} (\S~\ref{sec:det}) that locates and classifies all map elements by predicting element keypoints~$\bm{\mathcal{A}}=\{\bm{A}_i\in\mathbb{R}^{k\times 2}|i=1,\dots,N\}$ and their class labels~$\bm{\mathcal{L}}=\{l_i\in\mathbb{Z}|i=1,\dots,N\}$. 
The definition of element keypoint representation $\mathcal{A}$ is described in \S~\ref{sec:det}.
(3) A \textit{polyline generator}~(\S~\ref{sec:gen}) that produces a sequence of ordered polyline vertices which describes the local geometry of each detected map element $(\bm{A}_i, l_i)$.
An overview of three components is demonstrated in Figure~\ref{fig:overview}.

\begin{figure*}[t]
    \centering
    \includegraphics[width=\linewidth]{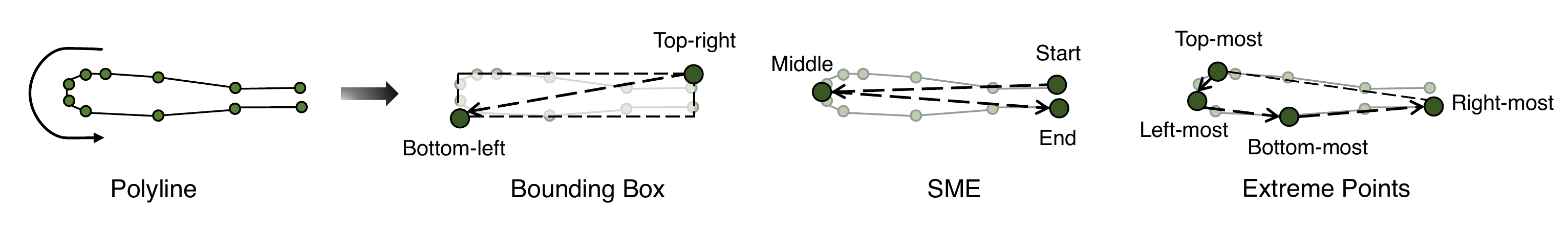}
    \vspace{-1em}
    \caption{Three different keypoint representations are proposed here: Bounding Box (k=2), SME (k=3), and Extreme Points (k=4), where $k$ has the same definition in \S~\ref{sec:VectorMapNet}: the number of key points of each keypoint representation. The arrow line indicates the direction of the example polyline, and the arrow dash lines indicate the vertices order of keypoint representations.
    }
    \label{fig:anchor}
    \vspace{-1em}
\end{figure*}

\subsection{BEV Feature Extractor}
\label{sec:bev_backbone}

The objective of BEV feature extractor is to lift various modality inputs into a canonical feature space and aggregates and align features these features into a canonical representation termed BEV features $\bm{\mathcal{F}}_{\mathrm{BEV}}\in\mathbb{R}^{W \times H \times (C_1+C_2)}$ based on their coordinates, where $W$ and $H$ represent the width and height of the BEV feature, respectively; $C_1$ and $C_2$ represent the output channels of the BEV feature extracted from the two common modalities: surrounding camera images $\mathcal{I}$ and LiDAR points $\mathcal{P}$. 

\noindent\textbf{Camera branch.}
We use ResNet to extract features from images, followed by a feature transformation module from image space to BEV space. VectorMapNet does not rely on certain feature transformation approaches and we opt to use a simple but popular variant of IPM, which produces BEV features of $\bm{\mathcal{F}}_{\mathrm{BEV}}^\mathcal{I}\in\mathbb{R}^{W \times H \times C_1}$. The detailed structure of the image extractor can be found in Appendix~\ref{model_details}.

\noindent\textbf{LiDAR branch.}
For LiDAR data $\mathcal{P}$, we use a variant of PointPillars~\citep{lang2019pointpillars} with dynamic voxelization~\citep{zhou2020end}, which divides the 3D space into multiple pillars and uses pillar-wise point clouds to learn pillar-wise feature maps.  We denote this feature map in BEV as $\bm{\mathcal{F}}_{\mathrm{BEV}}^\mathcal{P}\in\mathbb{R}^{W \times H \times C_2}$. 

For sensor fusion, we obtain the BEV features  $\bm{\mathcal{F}}_{\mathrm{BEV}}\in\mathbb{R}^{W \times H \times (C_1+C_2)}$ by concatenating $\bm{\mathcal{F}}_{\mathrm{BEV}}^\mathcal{I}$ and $\bm{\mathcal{F}}_{\mathrm{BEV}}^\mathcal{P}$, and then process the concatenated result with a two-layer convolutional network. 
An overview of the BEV feature extractor is shown at the bottom-left of Figure~\ref{fig:overview}.

\subsection{Map Element Detector}
\label{sec:det}
After extracting the bird's-eye view (BEV) features, VectorMapNet have to identify and abstractly represent map elements using these features. We employ a hierarchical representation for this purpose, specifically through element queries and keypoint queries, enabling us to model the non-local shape of map elements effectively. We leverage a variant of transformer set prediction detector~\citep{carion2020end} to achieve this goal, as it is a robust detector that eliminates the need for extra post-processing. Specifically, the detector represents map elements' locations and categories by predicting their element keypoints $\bm{\mathcal{A}}$ and class labels $\bm{\mathcal{L}}$ from the BEV features $\bm{\mathcal{F}}_{\mathrm{BEV}}$. 

\noindent\textbf{Element queries.}
The detector uses learnable element queries ${\bm{q}_{i}^{\mathrm{elem}}\in\mathbb{R}^{k\times d}| i=1,\dots,N_{\mathrm{max}}}$ as its inputs, where $d$ represents the hidden embedding size and $N_{\mathrm{max}}$ is a preset constant, which is much greater than the number of map elements~$N$ in the scene. The $i$-th element query $\bm{q}_{i}^{\mathrm{elem}}$ is composed of $k$ element keypoint embeddings $\bm{q}^{\mathrm{kp}}_{i,j}$: $\bm{q}^{\mathrm{elem}}_{i} = \{\bm{q}^{\mathrm{kp}}_{i,j}\in\mathbb{R}^d | j=1,\dots,k\}$.
Element queries are similar to object queries used in Detection Transformer (DETR)~\citep{carion2020end}, where a query represents an object. In our case, an element query represents a map element.

\noindent\textbf{Keypoint representations.} 
In object detection problems, people use bounding box to abstract object shape. Here we use $k$ element keypoints locations $\bm{A}_i = \{\bm{a}_{i,j}\in\mathbb{R}^2|j=1,...,k\}$ (please refer to Figure~\ref{fig:anchor}), to represent the outline of a map element. However, defining keypoints for map elements is not straightforward due to their diversity. We conduct an ablation study to investigate the performance of different choices in \S~\ref{subsec:anchor}. 
Note that element keypoints are different from polyline vertices and the element keypoints are intermediate representations of VectorMapNet that are passed to the polyline generator~(\S~\ref{sec:gen}) for conditional prediction, and the number of keypoints for each type of polyline is fixed and determined by its definition. Polylines are our output representations.

\noindent\textbf{Architecture.}
The overall architecture of the map element detector consists of a transformer decoder~\citep{vaswani2017attention} and a prediction head, as shown at the bottom-middle of Figure~\ref{fig:overview}. 
The decoder transforms the element queries using multi-head self-/cross-attention mechanisms. 
In particular, we use the deformable attention module~\citep{zhu2020deformable} as the decoder's cross attention module, where each element query has a 2D location grounding. It improves interpretability and accelerates training convergence~\citep{li2022dn}.

The prediction head has two MLPs, which decodes element queries into element keypoints $ \bm{a}_{i,j}=\mathrm{MLP_{kp}}(\bm{q}^{\mathrm{kp}}_{i,j})$ and their class labels $l_i = \mathrm{MLP_{cls}}([\bm{q}^{\mathrm{kp}}_{i,1}, \dots, \bm{q}^{\mathrm{kp}}_{i,k}])$, respectively. $[\cdot]$ is a concatenation operator.
Each keypoint embedding $\bm{q}^{\mathrm{kp}}_{i,j}$ in the map element detector consists of two learnable parts.
The first parts is a keypoint position embedding $\{\bm{e^{\mathrm{kp}}}_{j}\in\mathbb{R}^d| j=1,\dots,k \}$, indicating which position in an element keypoint the point belongs to. The second embedding $\{\bm{e^{\mathrm{p}}}_{i}\in\mathbb{R}^d | i=1,\dots,N_{\mathrm{max}} \}$ encodes which map element the keypoint belongs to. The keypoint embedding $\bm{q}^{\mathrm{kp}}_{i,j}$ is the addition of these two embeddings $\bm{e^{\mathrm{p}}}_{i}+\bm{e^{\mathrm{kp}}}_{j}$.

\begin{figure*}[htp]
    \centering
    \includegraphics[width=1.\linewidth]{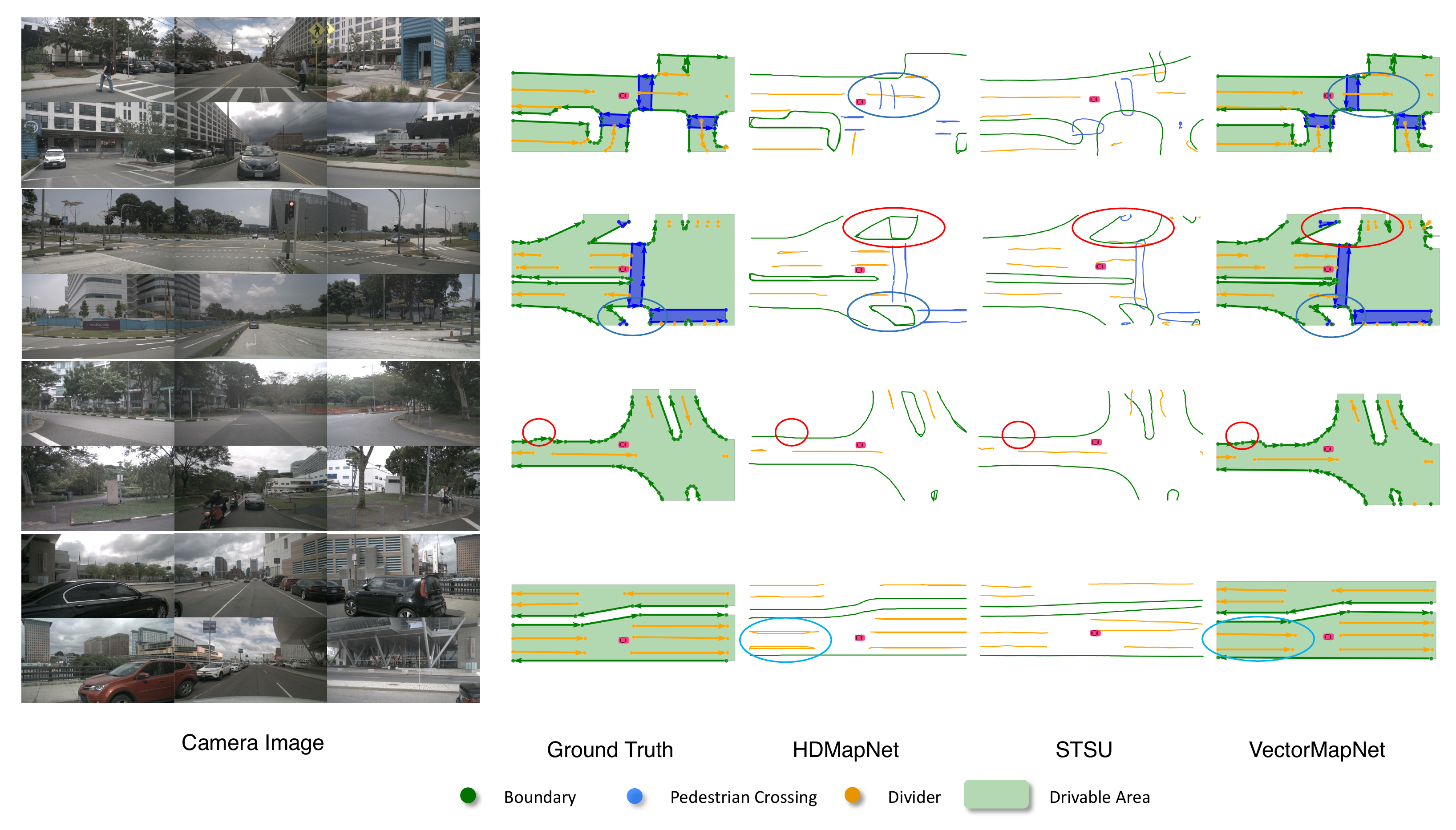}
    \vspace{-1em}
    \caption{
    Qualitative results generated by VectorMapNet and baselines. We use camera images as inputs for comparisons. The areas enclosed by \textbf{\textcolor{red}{red}} and \textbf{\textcolor{blue}{blue}} ellipses show that VectorMapNet can preserve sharp corners, and polyline representations prevent VectorMapNet from generating ambiguous self-looping results. The drivable area is inferred from disjoint boundaries. 
    }
    \label{fig:qualitative_HDMapNet} 
    \vspace{-1.em}

\end{figure*}

\subsection{Polyline Generator}
\label{sec:gen}
Upon the approximate position, shape, and category of map elements identified by map element detector, the polyline generator focuses on the detailed geometry of HD map, which entails calculating variable-length polyline vertices and their order. Accurate modeling of vertex relationships is crucial - for instance, a white line between two vertices often signifies a line connection in the vectorized map.
The polyline generator operates as a discrete distribution $p(\bm{V}_{i}^{\mathrm{poly}}| \bm{A}_{i}, l_{i}, \bm{\mathcal{F}}_{\mathrm{BEV}}^f)$ over the vertices of each polyline, conditioned on the initial layout~(\ie element keypoints $\bm{A}_{i}$ and class label $l_{i}$) and BEV features. To estimate this distribution, we decompose the joint distribution over each polyline  $\bm{V}_{i}^{\mathrm{poly}}$ as a product of a series of conditional vertex coordinate distributions. 
In particular, we transform each polyline $\bm{V}_i^{\mathrm{poly}}=\{\bm{v}_{i,n}\in\mathbb{R}^2 | n = 1,\dots,N_v\}$ into a flattened sequence $\{v_{i,n}^{f}\in\mathbb{R}|n = 1,\dots,2N_v\}$ by concatenating coordinates values of polyline vertices and add an additional \textit{End of Sequence} token~($EOS$) at the end of each sequence, and the target distribution turns into:
\begin{equation}
    \resizebox{0.9\hsize}{!}{%
    $p(\bm{V}_{i}^{\mathrm{poly}}| \bm{A}_{i}, l_{i}, \bm{\mathcal{F}}_{\mathrm{BEV}}; \bm{\theta}) = \prod_{n=1}^{2N_v} p(v_{i,n}^{f}| v_{i,<n}^{f}, \bm{A}_{i}, l_{i}, \bm{\mathcal{F}}_{\mathrm{BEV}}).$
    }
\end{equation}
Following PolyGen~\cite{nash2020polygen}, we use a categorical distribution to model the probability of each vertex position given the preceding vertex position. This allows us to model the complex and irregular shapes of map elements while maintaining the efficiency of discrete distributions. And we model this distribution using an autoregressive network that outputs the parameters of a predictive distribution at each step for the next vertex coordinate. This predictive distribution is defined over all possible discrete vertex coordinate values and $EOS$. 

\noindent\textbf{Vertices as discrete variables.} 
Using discrete distributions to model polyline vertices has the advantage of representing arbitrary shapes, \ie categorical distributions can easily represent various polylines, such as multi-modal, skewed, peaked, or long-tailed, that are commonly seen in our task. 
Thus, we quantize the coordinate values into discrete tokens and model each token with a categorical distribution.
We also conduct an ablation study in Appendix~\S~\ref{subsec:vertexmodeling} to investigate other choices.

\noindent\textbf{Architecture.}
To model these local geometric structures of polylines, the autoregressive network we choose is Transformer~\citep{vaswani2017attention}~(see the bottom-right of Figure~\ref{fig:overview}). Transformer architecture has consistently demonstrated superior performance in conditional sequence generation tasks and are highly effective at capturing the vertex dependencies present in map data. Each polyline's keypoint coordinates and class label are tokenized and fed in as the query inputs of the transformer decoder. Then a sequence of vertex tokens are fed into the transformer iteratively, integrating BEV features with cross-attention, and decoded as polyline vertices. Note that the generator can generate all polylines in parallel. 

\noindent\textbf{Vertex embeddings.} 
Following PolyGen~\citep{nash2020polygen}, we use an addition of three learned embeddings as the embedding of each vertex token: \textit{Coordinate Embedding}, indicating whether the token represents $x$ or $y$ coordinate; \textit{Position Embedding}, representing which vertex the token belongs to; \textit{Value Embedding}, expressing the token's quantized coordinate value.

\subsection{Learning}
\label{subsec:learning}
We train our model by minimizing the sum of map element detector loss and polyline generator loss:
\begin{equation}
    \mathcal{L} = \mathcal{L}_{det} + \mathcal{L}_{gen}.
    \label{eq:total_loss}
\end{equation}

\noindent\textbf{Map element detector loss.} 
 Following \citep{wang2022detr3d, zhu2020deformable}, the detector is trained with bipartite matching loss, thus avoiding post-processing steps like non-maximum suppression (NMS). We describe the detail of the map element detector loss~$\mathcal{L}_{det}$ function in Appendix \S~\ref{subsec:loss}.
 
\noindent\textbf{Polyline generator loss.} 
Polyline generator is trained to maximize the log-probability of the polyline vertices. We use negative log-likelihood as its loss function: 
\begin{equation}
\mathcal{L}_{gen}=- \frac{1}{2N_v} \sum_{n=1}^{2N_v} \log \hat{p}(v_{i,n}^{f}| v_{i,<n}^{f}, \bm{A}_{i}, l_{i}, \bm{\mathcal{F}}_{\mathrm{BEV}}^f),
\end{equation}
where $\hat{p}(v_{i,n}^{f}|\dots)$ is the conditional probability of discrete coordinate value $v_{i,n}^{f}$, and $v_{i,<n}^{f}$ are ground truth discrete coordinate values with index less than $n$.
The default training strategy is teacher forcing, meaning that we use ground truth keypoints as generator input. To avoid the exposure bias~\citep{bengio2015scheduled}, we further experiment with first training with teacher forcing, and then fine-tuning with predicted keypoints.

\section{Experiments}
\label{sec:experiment}
\noindent\textbf{Experiments protocol.} We conduct experiments on the nuScenes~\citep{caesar2020nuscenes} and Argoverse2~\citep{Argoverse2} dataset. 
Following HDMapNet~\citep{li2021hdmapnet}, we assess the quality of a predicted HD map by comparing its components (\ie polylines) with ground truth.
Both HDMapNet and our paper use Chamfer distance for polyline matching (Chamfer AP). Additionally, we also introduced another distance metric termed Fr\'echet distance (Fr\'echet AP), which better measures the distance between polylines by considering the order of vertices. The definitions and calculation processes of Chamfer AP and Fr\'echet AP are in \S~\ref{sec:metric}.
Additionally, the details of dataset settings~(\S~\ref{subsec:dataset}), implementations~(\S~\ref{sec:implementation}), and additional qualitative results~(\S~\ref{sec:additional_qualitative}) are presented in the Appendix as well. 

\begin{table*}[] 
    \centering
    \caption{Results on nuScenes dataset. Fusion denotes the model using both images and LiDAR points as inputs.
    Methods with fine-tune means the model is applied two stage training strategy introduced in \S~\ref{subsec:learning}
    }
    \vspace{-1.em}
    \scalebox{0.85}{
    \begin{tabular}{@{}lllll@{}}
    \toprule
    \multicolumn{1}{c}{Methods} &AP$_{ped}$ & AP$_{divider}$ & AP$_{boundary}$ & mAP \\ \midrule
    STSU~\citep{can2021structured} &7.0  &11.6  &16.5  &11.7\\
    HDMapNet (Camera)~\citep{li2021hdmapnet} &14.4 & 21.7 &33.0  &23.0 \\
    HDMapNet (LiDAR)~\citep{li2021hdmapnet} &10.4  &24.1  &37.9  &24.1 \\
    HDMapNet (Fusion)~\citep{li2021hdmapnet} &16.3  & 29.6 &46.7 & 31.0 \\
    \hline
    VectorMapNet (Camera) &36.1 &47.3 &39.3 &40.9 \\
    VectorMapNet (Camera) + fine-tune &42.5 &51.4 &44.1 &46.0 \\
    \hline
    VectorMapNet (LiDAR) &25.7 &37.6 &38.6 &34.0  \\
    VectorMapNet (Fusion) &37.6 &50.5 &47.5 &45.2 \\
    VectorMapNet (Fusion) + fine-tune &\textbf{48.2} &\textbf{60.1} &\textbf{53.0} &\textbf{53.7} \\ \bottomrule

    \end{tabular}
    }
    \label{tab:baseline}
    \vspace{-0.5em}

\end{table*}
\begin{table*}[!ht] \centering
    \ra{1.2}
    \caption{ 
        Results on Argoverse2 dataset.
    }
    \vspace{-1em}
    \resizebox{0.85\textwidth}{!}{
    \begin{tabular}{@{}lc|llllcllll@{}}
    \toprule
    \multicolumn{2}{c}{}  & \multicolumn{4}{c}{Fr\'echet Distance} && \multicolumn{4}{c}{Chamfer Distance}\\ \cmidrule(lr){3-6} \cmidrule(lr){8-11}
    Keypoint Representaion & $\#dim$ & AP$_{ped}$ & AP$_{divider}$ & AP$_{boundary}$ & mAP && AP$_{ped}$ & AP$_{divider}$ & AP$_{boundary}$ & mAP \\ \hline

    HDMapNet (Camera)~\cite{li2021hdmapnet} &2 &- &- &-  &-             &&13.1 &5.7 &37.6 &18.8\\
    VectorMapNet (Camera)                   &2 &43.2 &45.5 &52.0 &46.9  &&38.3 &36.1 &39.2 &37.9 \\ \hline
    VectorMapNet (Camera)                   &3 &41.7 &42.3 &49.9 &44.6  &&36.5 &35.0 &36.2 &35.8 \\ 
    \bottomrule
    \end{tabular}
    }
    \label{tab:argoverse2}
    \vspace{-1em}
\end{table*}

\subsection{Comparison with Baselines}
\label{sec:baseline}
The HD semantic map construction is a new problem, and there are no established methods to compare with. Therefore, we carefully chose two baselines HDMapNet and STSU~\citep{can2021structured} that are representative and can effectively compare with VectorMapNet. Specifically: HDMapNet can provide valuable insights into the effectiveness of commonly used map segmentation methods for HD semantic map construction. STSU, a direct map structure learning method, can provide valuable insights into its effectiveness for HD semantic map construction. Moreover, our baseline comparison also includes the results of HDMapNet and VectorMapNet using different modalities as inputs, which demonstrate the impact of different feature extraction methods on HD semantic map construction. The details of baselines model are described in Appendix~\S~\ref{sec: baseline model settings}. We report the average precision that uses Chamfer distance as the threshold to determine the positive matches with ground truth. $\{ 0.5, 1.0, 1.5 \}$ are the predefined thresholds of Chamfer distance AP. 

\noindent\textbf{Results on nuScenes.} As shown in Table~\ref{tab:baseline}, VectorMapNet outperforms HDMapNet by a large margin under all settings (+17.9~mAP in Camera, +9.9~mAP in LiDAR, and +14.2~mAP in Fusion). Compared to camera-only and LiDAR-only, sensor fusion introduces +4.3~mAP improvement and +11.2~mAP improvement, respectively.
As described in \S~\ref{subsec:learning}, our two stage training strategy further boosts the performance of both camera-only and sensor fusion methods by +6.9 mAP and +8.5 mAP, respectively.
STSU is -29.2~mAP lower than VectorMapNet. Since STSU treats all map elements as a set of fixed-size segments, we hypothesize that ignoring the fine geometry of map elements hurts the performance. 

\noindent\textbf{Results on Argoverse2.} We further compare HDMapNet and VectorMapNet on Argoverse2 dataset, shown in Table~\ref{tab:argoverse2}.
Since Argoverse2 provides z-axis annotations, we give VectorMapNet results both in 2D and 3D.
In many cases of Argoverse2, the annotated boundaries and divider lines overlap with each other, making it difficult for models to separate them. It results in a drop in performance of both methods, especially in AP$_{divider}$ of HDMapNet (21.7 AP$_{divider}$ to 5.7 AP$_{divider}$) because its rasterized representation fails to handle these cases. In contrast, VectorMapNet remains competent, showing the advantage of using vectorized representation to represent overlapping elements.

\begin{figure}[t]
  \centering
  \includegraphics[width=1.\linewidth]{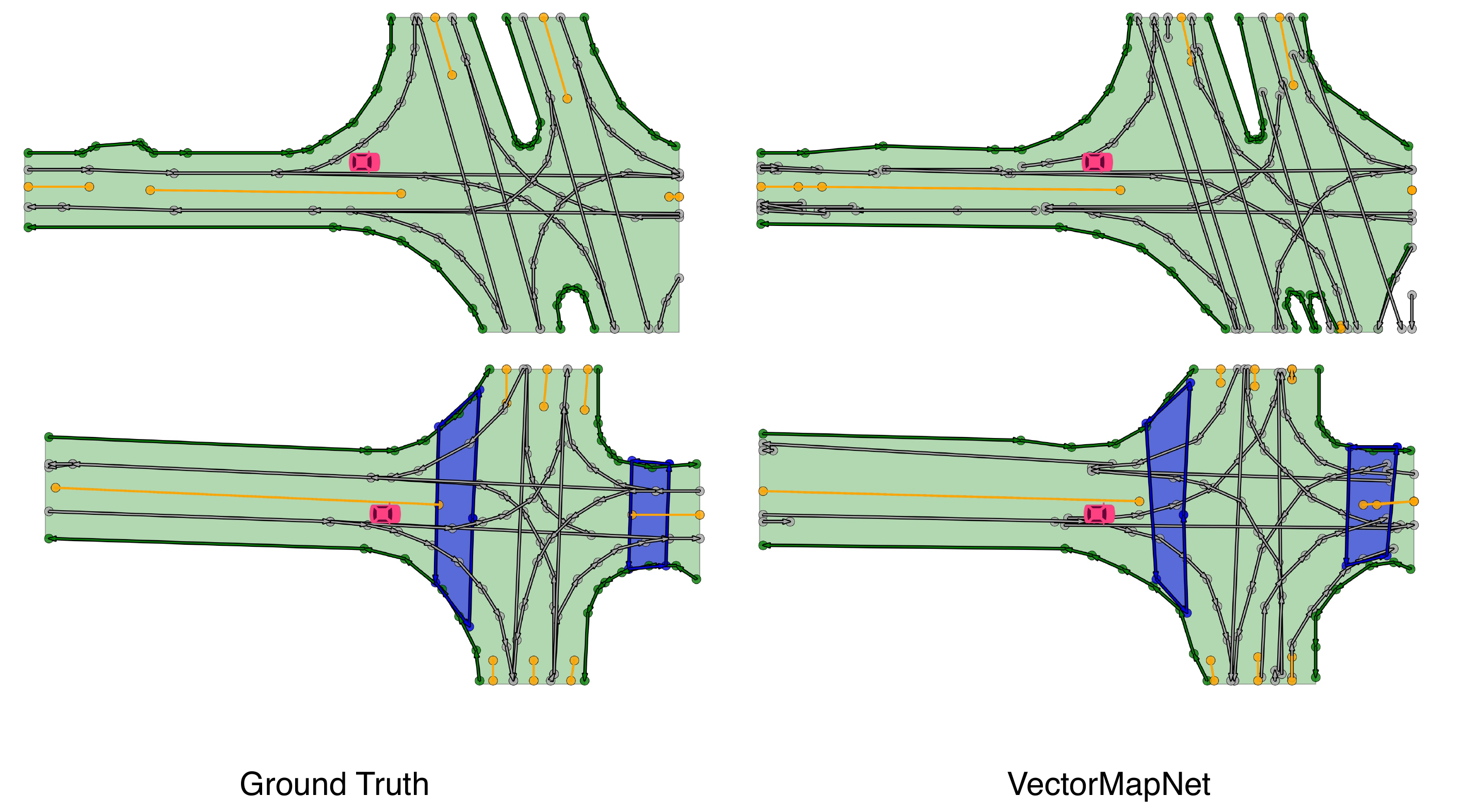}
  \caption{The centerline predictions by VectorMapNet, where the gray lines are the predicted centerlines.}
  \vspace{-1em}
  \label{fig:centerline_prediction}
\end{figure}

\vspace{-1.em}
\subsection{Qualitative Analysis}
\begin{figure}[!ht]
  \centering
  \includegraphics[width=1.\linewidth]{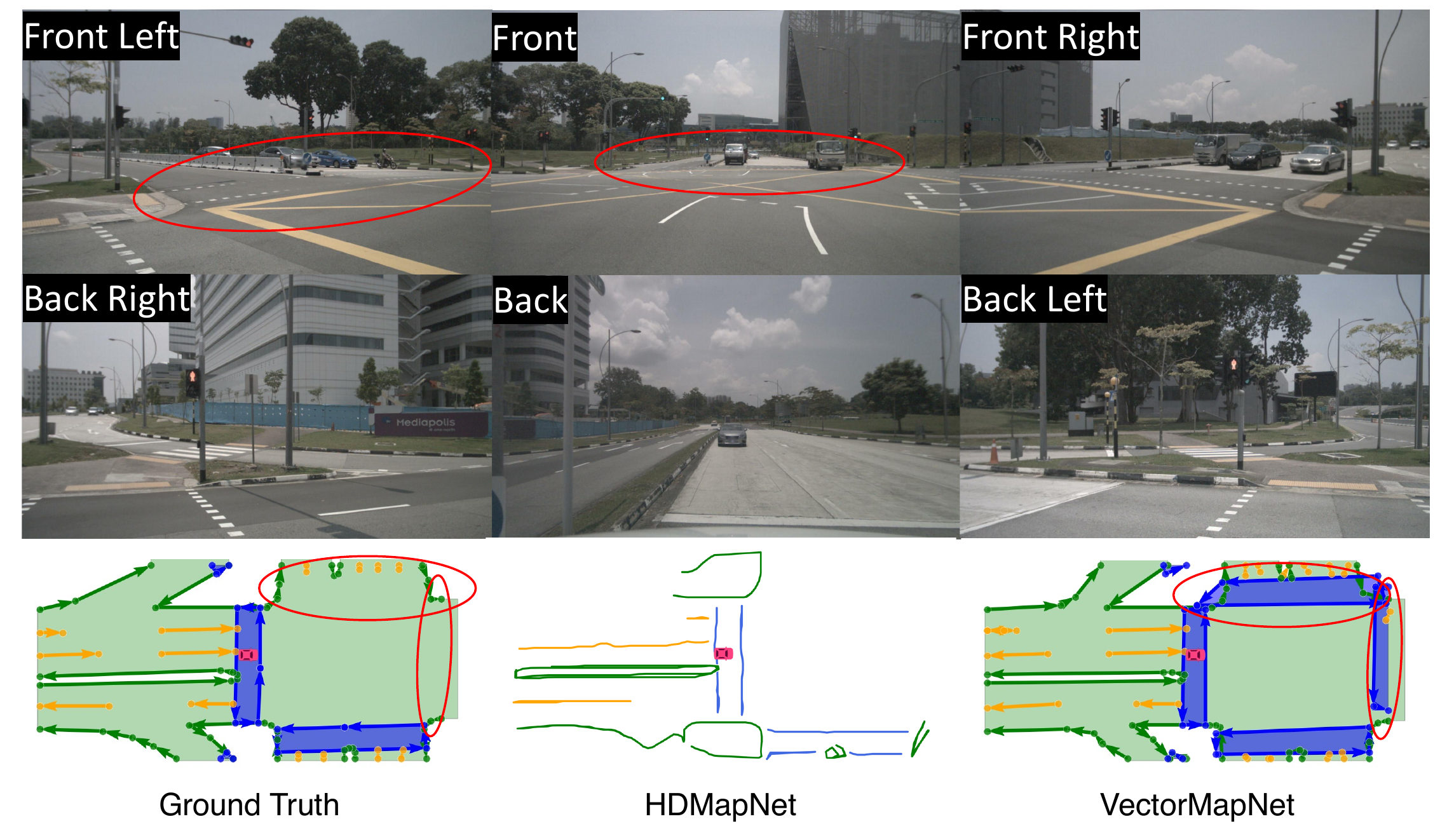}
  \caption{An example of VectorMapNet detecting unlabeled map elements. The \textbf{\textcolor{red}{red ellipses}} indicate two pedestrian crossings that are missing in ground truth annotations, while VectorMapNet detects it correctly. All the predictions are generated from camera images.}
  \label{fig:qualitative_example}
  \vspace{-1em}
\end{figure}

\noindent\textbf{Benefits of using polylines as primitives.}
From visualizations, we find that using polylines as primitives has brought us two benefits compared with baselines: 
First, polylines effectively encode the detailed geometries of map elements, \eg the corners of boundaries (see the red ellipses in Figure~\ref{fig:qualitative_HDMapNet}). 
Second, polyline representations prevent VectorMapNet from generating ambiguous results, as it consistently encodes direction information. In contrast, Rasterized methods are prone to falsely generating loopy curves (see the blue ellipses in Figure~\ref{fig:qualitative_HDMapNet}). These ambiguities hinder safe autonomous driving. 
Therefore, the polyline is a desired primitive for map learning, as it can reflect real-world road layouts and explicitly encode directions.

\noindent\textbf{Benefits of posing map learning as a detection problem.}
VectorMapNet operates in a top-down detection manner: it first models the map's topology and the locations of map elements, then generates the details of these elements. Visualizations demonstrate that VectorMapNet captures all map elements comprehensively, even the smaller ones near edges. The high mAP of VectorMapNet, when compared to other baselines, validates this observation. We attribute these impressive results to the model's ability to model topological relationships between map elements, thus implicitly capturing complex scene interrelationships. This is evidenced by Figure~\ref{fig:qualitative_example}, where the model identifies pedestrian crossings at intersections that are missed in the annotations of the HD map provided by the dataset. Although these relationships are not explicitly taught, the model learns them via controlled information propagation between query embeddings, using self-attention modules — a technique from the original Transformer paper. This showcases the model's proficient scene understanding.

\noindent\textbf{Centerline prediction by VectorMapNet.}
As discussed in \S~\ref{sec:model_overview} and above, the polyline is a versatile primitive, capable of representing map element classes that extend beyond the elements in the HD semantic map setting. To further demonstrate this flexibility, we expand VectorMapNet to predict the centerline, an imaginary line commonly used as a reference for driving direction, vehicle positioning, and navigation. The adaptation is quite straightforward: VectorMapNet treats centerlines as a set of polylines and implicitly encodes their topological relations. This process involves no modifications to the model structure. Figure~\ref{fig:centerline_prediction} displays the results of VectorMapNet's centerline prediction. 

\begin{table*}[] 
    \centering
    \ra{1.3}
    \caption{ 
        Ablation study of keypoint representaions. $k$ is the keypoint number of each keypoint representation. 
    }
    \resizebox{0.85\textwidth}{!}{
    \begin{tabular}{@{}cc|llllcllll@{}}
    \toprule
    \multicolumn{2}{c}{}  & \multicolumn{4}{c}{Fr\'echet Distance} && \multicolumn{4}{c}{Chamfer Distance}\\ \cmidrule(lr){3-6} \cmidrule(lr){8-11}
    Keypoint Representaion & $k$ & AP$_{ped}$ & AP$_{divider}$ & AP$_{boundary}$ & mAP && AP$_{ped}$ & AP$_{divider}$ & AP$_{boundary}$ & mAP \\ \hline
    Bbox &2     &\textbf{47.4} &46.9 &\textbf{62.8} &\textbf{52.4}  &&\textbf{36.1} &\textbf{47.3} &\textbf{39.3} &\textbf{40.9} \\
    SME  &3     &47.0 &\textbf{47.4} &56.9 &50.4  &&27.6 &34.4 &35.4 &32.5 \\
    Extreme &4  &41.7 &47.3 &59.0 &49.4  &&30.4 &33.1 &37.3 &33.6 \\
    \bottomrule
    \end{tabular}
    }
    \label{tab:anchor_ablation}
\end{table*}
\begin{table}[!ht] 
    \centering
    \caption{
    	The benefits of predicted maps in improving the motion forecasting baseline. There are three input settings: past trajectories (denoted as Traj.), past trajectories with the human-annotated HD map from the nuScenes (denoted as Traj. + G.T. Map), and past trajectories with the predicted map from VectorMapNet (denoted as Traj. + Pred. Map). The predicted map greatly improves the prediction performance compared with the model that only use past trajectories.
    }
    \scalebox{0.95}{
    \begin{tabular}{@{}llll@{}}
    \toprule
    \multicolumn{1}{c}{Model Inputs} & minADE $\downarrow$ & minFDE$\downarrow$ & MR@2m$\downarrow$ \\ \midrule
    Traj.             &0.909  &1.577  &19.6 \\
    Traj. + G.T. Map    &0.779  &1.390 &18.0   \\
    Traj. + Pred. Map &0.826  &1.477  &18.2 \\ \bottomrule
    
    \end{tabular}
    }
    \label{tab:prediction}

\end{table}

\subsection{Ablation Studies}
\label{subsec:ablation}
We provide ablation studies for keypoint representation in this section. For other ablation studies (\ie curve sampling strategies, vertex modeling methods, and extrinsic robustness), please refer to Appendix~\S~\ref{sec:more_ablations}.

\noindent\textbf{Keypoint representations.}
\label{subsec:anchor}
Since there is no straightforward keypoint design to represent map elements with few fixed number of points, we propose three simple representations as shown in Figure~\ref{fig:anchor}: \textit{Bounding Box~(Bbox)}, which is the smallest box enclosing a polyline, and its keypoints are defined as the top-right and bottom-left points of the box; \textit{Start-Middle-End (SME)}, which samples the start, middle, and end point from a polyline; \textit{Extreme Points}, which are the left-most, right-most, top-most, and bottom-most points of a polyline. 
We experiment with these representations and list the results in Table~\ref{tab:anchor_ablation}.
Our results show that the bounding box representation leads to the best mean average performance in both metrics, outperforming others by 2.0 Fr\'echet mAP and 7.3 Chamfer mAP.

\subsection{Motion Forecasting with Vectorized HD Maps from VectorMapNet}
To evaluate the capacity of our method to understand scene relationships and to investigate its usefulness in subsequent tasks, we put our predicted HD map to the test within a motion forecasting task. This task heavily relies on precise map information for accurate prediction of future motion.

\noindent\textbf{Task Settings.} The motion forecasting requires that the model have to predict 6 possible future trajectories (3 seconds) from past agents' trajectories (1 second) and an HD semantic map spanning $60m\times30m$. Data is generated from the nuScenes tracking dataset, selecting agents with complete 3-second future observations. This results in 25,645 training and 5,460 test samples. We examine three input scenarios: past trajectories alone, past trajectories with the true HD map, and past trajectories with the VectorMapNet predicted map. We utilize mmTransformer~\citep{liu2021multimodal} for motion forecasting due to its versatility in using map data or relying solely on past trajectories. This assists in assessing the quality of our learned maps.

\noindent\textbf{Results.} To evaluate the performance of motion forecasting under different input settings, we report results on three commonly used metrics~\citep{chang2019argoverse}: minimum average displacement error (minADE), minimum final displacement error (minFDE) and miss rate (MR). To get the results, these metrics only account for the best trajectory out of 6 predicted trajectories. Results in Table~\ref{tab:prediction} show that the map predicted by VectorMapNet has encoded environment information that greatly helps the motion forecaster, compared with the model that only takes past trajectories as inputs. The gap between the ground-truth map and the predicted map is not big either, especially in terms of MR (-0.2\%). We think future research could further close the performance gap.

\section{Discussions}
\noindent\textbf{Limitations.} It is worth noting that the model has some limitations, and we leave it for future works. \textit{Lacking Temporal Information}: The model generates coherent geometries in a single frame but doesn't guarantee temporally consistent predictions. \textit{Mismatch Problem of a Two-stage Model}: A feature space mismatch exists between the map element detector and the polyline generator due to the teacher-forcing training strategy. Although fine-tuning is necessary for optimal performance, it results in tricky training schedules. \textit{Hallucination Ability}: The model can make predictions at locations that are occluded and not visible to cameras, showcasing its scene understanding capabilities. However, this reduces the model's interpretability.

For further discussions, such as the potential societal impact of our method, please refer to Appendix~\S~\ref{sec:Discussions}.

\section{Conclusions}
We present VectorMapNet, an end-to-end model to tackle the HD semantic map learning problem. Unlike existing works, VectorMapNet uses polylines as the primitives to represent vectorized HD map elements. To predict polylines from sensor data, we decompose the problem into a detection step and a generation step. Our experiments show that VectorMapNet can generate coherent and complex geometries for urban map elements, benefiting from the polyline primitives. We believe that this novel way to learn HD maps provides a new perspective on the HD semantic map learning problem.

\section*{Acknowledgements}
This work is supported by the National Key R\&D Program of China (2022ZD0161700). 
We would like to thank Qi Li and Tianyuan Zhang for their help on various baselines, and thank Ziyuan Huang and Bowen Li for paper proofreading.

\clearpage




\bibliography{egbib}
\bibliographystyle{icml2023}

\newpage
\appendix
\onecolumn

\label{Appendix}

\section{Experiment Setup}
\subsection{Dataset} 
\label{subsec:dataset}
\noindent\textbf{nuScenes}
We experiment on nuScenes~\citep{caesar2020nuscenes} dataset, which contains 1000 sequences of recordings collected by autonomous driving cars. Each episode is annotated at 2Hz and contains 6 camera images and LiDAR sweeps. 
Our dataset setup and pre-processing steps are identical to that of HDMapNet~\citep{li2021hdmapnet}, which includes three categories of map elements -- pedestrian crossing, divider, and road boundary -- from the nuScenes dataset.

\noindent\textbf{Argoverse2}
We further conduct experiments on Argoverse2~\citep{Argoverse2} dataset. Like nuScenes, it contains 1000 logs (700, 150, 150 for training, validation and test set). Each episode provides 15s of 20Hz camera images, 10Hz LiDAR sweeps and a vectorized map. We use the same pre-processing settings as on nuScenes dataset.

\subsection{Metrics}
\label{sec:metric}
In contrast to existing methods which generate rasterized results, our method does not require rasterizing curves on grids. Therefore, we opt not to use Intersection-Over-Union (IoU) as a metric. We use a distance-based metric to evaluate the similarity between predicted curves and ground-truth curves. 
We follow the instance-level evaluation metric proposed by HDMapNet~\citep{li2021hdmapnet} to compare the instance-level detection performance of our model to baseline methods.
The metric is average precision (AP), where positive/negative samples are based on geometric similarity, more concretely, Chamfer distance and Fr\'echet distance. For clarity, we call the AP based on Chamfer distance and Fr\'echet distance as Chamfer AP and Fr\'echet AP, respectively.

\noindent\textbf{Chamfer distance.}
Chamfer distance is a distance measure that quantifies the similarity between two \emph{unordered} sets.
The Chamfer distance is an evaluation metric that quantifies the similarity between two unordered sets by taking into account the distance of each permutation of the elements of set as follows: 
\begin{equation}
    D_{chamfer}(\mathcal{S}_{1}, \mathcal{S}_{2}) = 
        \frac{1}{2}(
            \frac{1}{|\mathcal{S}_{1}|} 
                \sum_{p \in \mathcal{S}_{1}} \mathop{\min}_{q\in\mathcal{S}_{2}} \|p,q\|_2 + 
            \frac{1}{|\mathcal{S}_{2}|} 
                \sum_{q \in \mathcal{S}_{2}} \mathop{\min}_{p\in\mathcal{S}_{1}} \|q,p\|_2) .
\end{equation}

In our experiments, we use chamfer distance to calculate the distance between a prediction and a ground truth polyline set, and each polyline set is represented by uniformly sampling a polyline to $N_{pts}$ vertices, where $N_{pts}$ is set to $100$ in our experiments.

\noindent\textbf{Fr\'echet distance.}
The order of polyline vertices is not measured by Chamfer distance. Therefore, we introduce Fr\'echet distance as an additional measure. Fr\'echet distance is a measure of similarity of curves that takes both the positions and the \emph{order} of the points along the curves into consideration. Our implementation is based on discrete Fr\'echet distance~\citep{eiter1994computing,agarwal2014computing}. 

We use the discrete version of Fr\'echet distance~\citep{eiter1994computing,agarwal2014computing} to evaluate the geometric similarity between two polyline $P$ and $Q$. We denote $\sigma(P)$ as a sequence of endpoints of the line segments of $P$. In particular, $\sigma(P)= (p_{1}, \dots, p_{m})$ is a sequence with $m$ vertices that uniformly sampled from the original input polyline $P$, where 
each position of $P$ between $p_i$ and $p_{i+1}$ can be approximated by using an affine transformation that is $p_{i + \lambda} = (1 - \lambda)p_{i} + \lambda p_{i + 1}$ and the $m$ in our experiment is set as $100$.

Let $P$ and $Q$ be polyline and $\sigma(P) = (u_{1}, \dots, u_{p})$ and $\sigma(Q) = (v_{1}, \dots, v_{q})$ the corresponding sequences.
A coupling $L$ is a sequence of distinct pairs between $\sigma(P)$ and $\sigma(Q)$: 
\begin{equation}
(u_{a_1}, v_{b_1}), \dots ,(u_{a_m}, v_{b_m}).
\end{equation}
These indexes $\{a_1,\dots,a_m\}$ and $\{b_1,\dots,b_m\}$ are nondecreasing surjection such that $a_1 = 1$, $a_m = p$, $b_1 = 1$, $b_m = q$ and for all $ i<j \in \{1, \dots, q\}$, $a_{i}\leq a_{j}$ and $b_{i}\leq b_{j}$.

We define the norm $\| L \|$ of the $L$ is the length of the longest pair in $L$, that is,
\begin{equation}
    \| L \| = \underset{i=1,\dots,m}{\max} d(u_{a_i}, v_{b_i}).
\end{equation}

The discrete Fr\'echet distance between polyline $P$ and $Q$ is defined to be
\begin{equation}
    \delta_{dF}(P, Q) = \min \{ \| L \|, \text{$L$ is a coupling between $P$ and $Q$} \}.
\end{equation}
This equation indicates that the distance of discrete Fr\'echet distance is the minimum norm of all possible couplings. To Find the coupling plausible $L$ that has the minimum norm, we use a Dynamic programming-based algorithm that is described in Algorithm~\ref{alg:discrete}.

\begin{algorithm}[H]
    \caption{The Algorithm of Discrete Fr\'echet Distance}
    \SetKwProg{Fn}{Function}{}{end}
    \SetKwFunction{FMain}{$c$}
    \SetAlgoLined
    \KwIn{polyline $P = (u_1, \dots , u_p)$ and $Q = (v_1, \dots, v_q)$.}
    \KwOut{$\delta_{dF}(P, Q)$}
    $ca$ : an 2d array of real with size of $(p \times q)$\;
    \Fn{\FMain{$i$, $j$}}{
        \uIf{$ca(i, j) > -1$}{
            \KwRet $ca(i, j)$\;
        }
        \uElseIf{$i = 1$ and $j = 1$}{
            $ca(i, j) := d(u1, v1)$\;
        }
        \uElseIf{$i > 1$ and $j = 1$}{
            $ca(i, j) := \max\{ c(i - 1, 1), d(u_i, v_1) \}$\;
        }
        \uElseIf{$i = 1$ and $j > 1$}{
            $ca(i, j) := \max \{ c(1, j - 1), \,d(u_1, v_j) \}$\;
        }
        \uElseIf{$i > 1$ and $j > 1$}{
            $ca(i, j) := \max\{ \min( c(i - 1, j), c(i - 1, j - 1), c(i, j-1)), d(u_i,v_j)\}$\;
        }
        \Else{
            $ca(i, j) := \infty$ \;
        }
        \KwRet $ca(i, j)$\;
    }
    \Begin{
    \For{$i = 1$ to $p$}{
        \For{$j = 1$ to $q$}{
            ca(i, j) := -1.0;
        }
    }
    \Return{$c(p, q)$\;}
    }
    
    \label{alg:discrete}
\end{algorithm}

\begin{figure}[t]
  \centering
  \includegraphics[width=1.\linewidth]{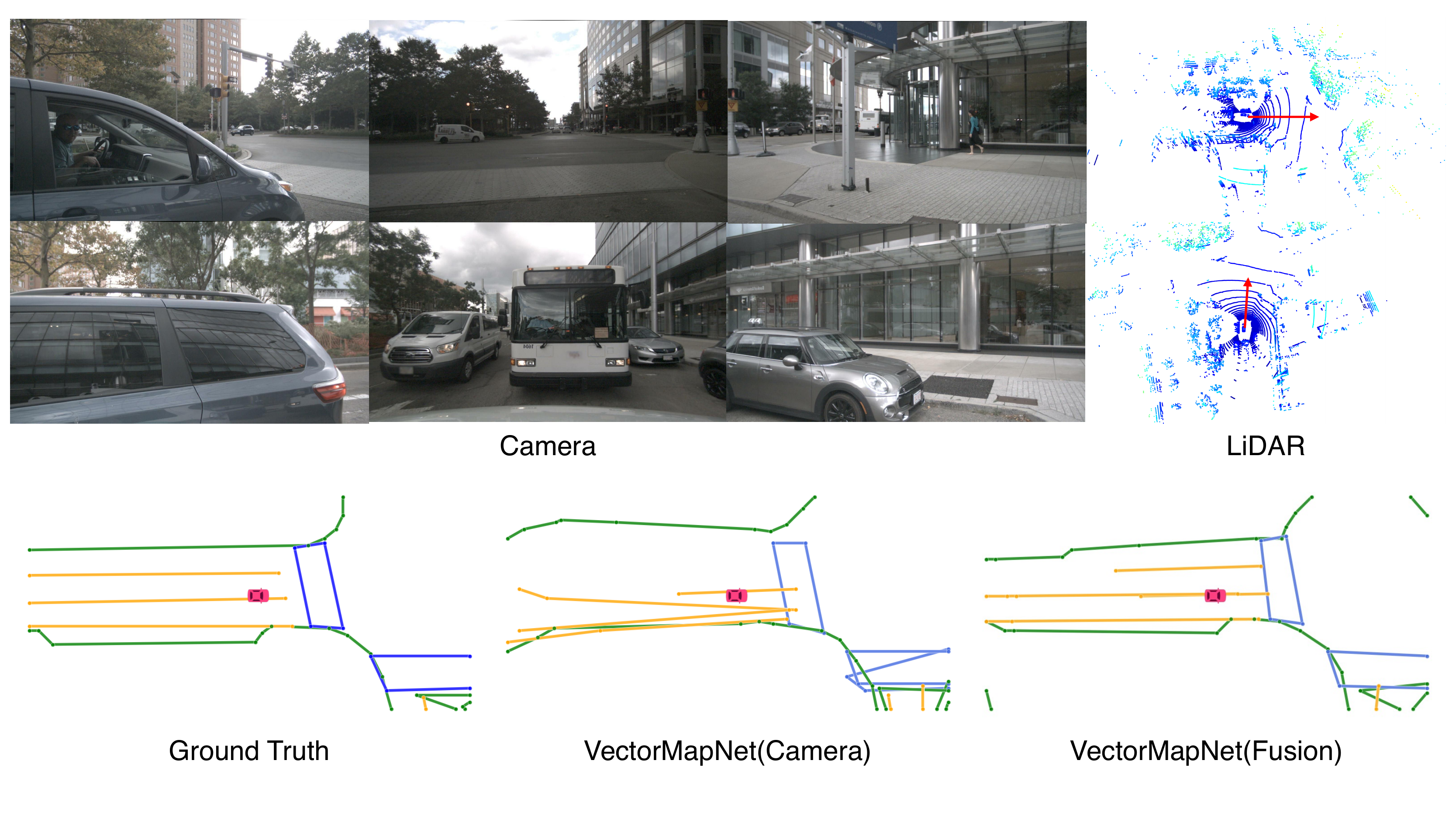}
  \caption{When the ego car cameras are occluded by the nearby vehicles, VectorMapNet(Camera) can not precept the surrounding map. With the depth cue from LiDAR, VectorMapNet(Fusion) can generate a more plausible result than its camera counterpart.}
  \label{fig:qualitative_Fusion1}
\end{figure}

\begin{figure}[t]
  \centering
  \includegraphics[width=1.\linewidth]{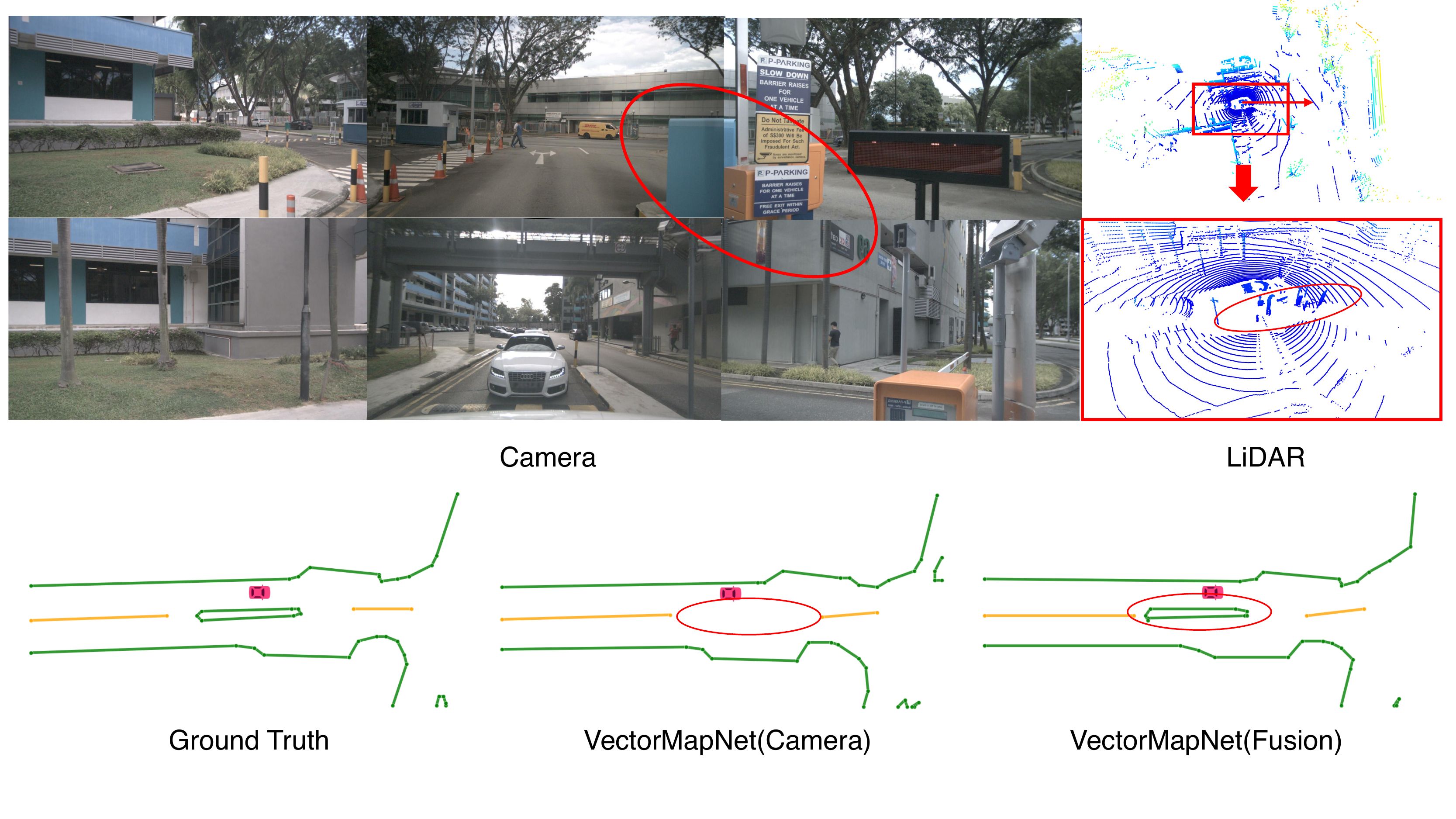}
  \caption{The blind area of onboard cameras may cause our model to miss the map elements closed ego vehicle. In contrast, we can easily find that LiDAR data has sensed some obstacles near the ego vehicle in the right-most column. With these cues, our fusion model detects the missed lane boundary by our camera-only model.}
  \label{fig:qualitative_Fusion2}
\end{figure}

\begin{figure}[t]
  \centering
  \includegraphics[width=1.\linewidth]{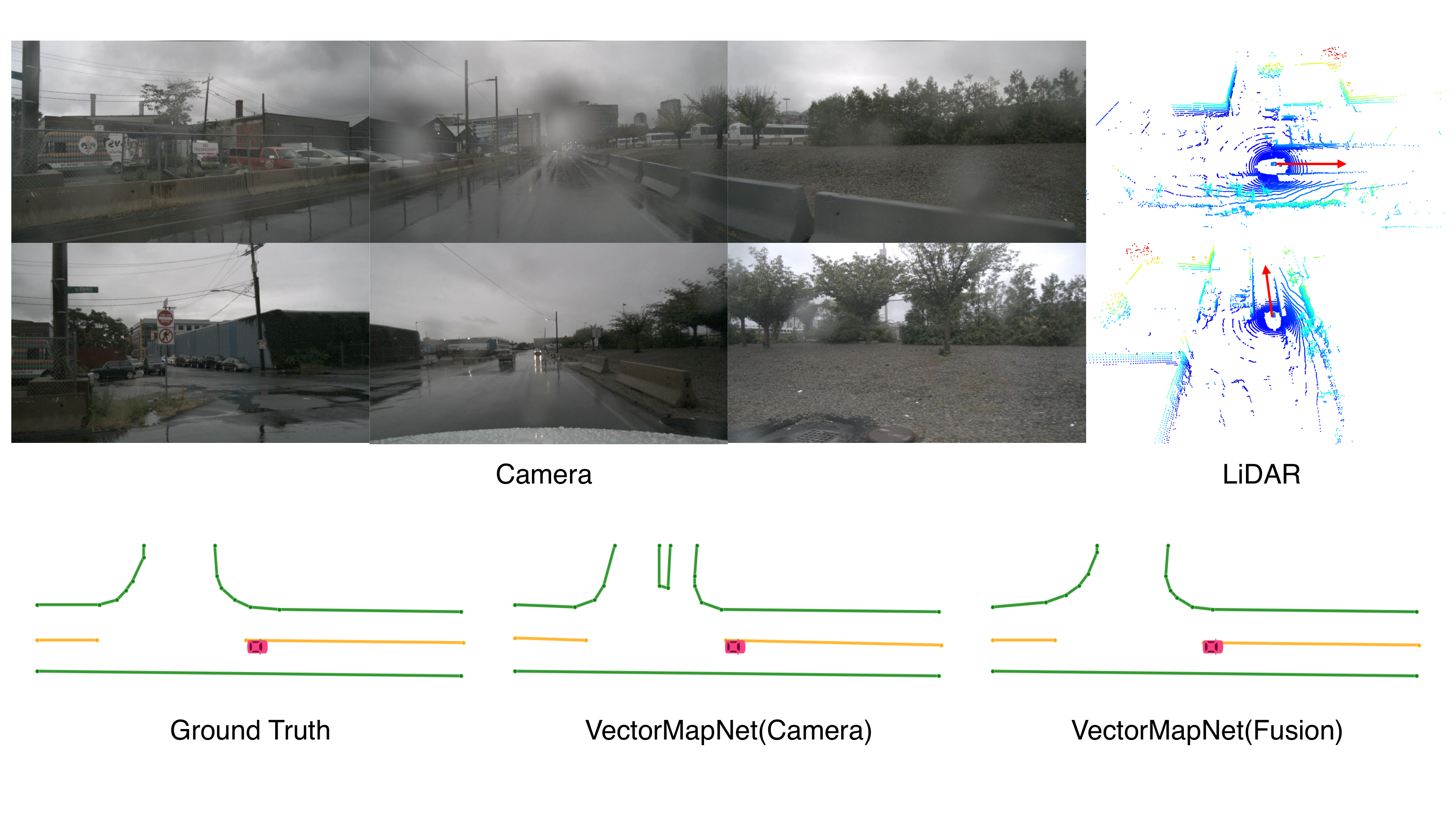}
  \caption{The qualitative results of VectorMapNet in bad weather conditions. VectorMapNet(Camera) falsely detects these puddles near the intersection as a lane boundary. The fusion result shows that the miss detection issue can be resolved by combining the depth information.}
  \label{fig:qualitative_Fusion3}
\end{figure}

\section{More Qualitative results of VectorMapNet}
\label{sec:additional_qualitative}

\subsection{Visualization results of VectorMapNet(Fusion)}
We visualized three cases of VectorMapNet (Fusion) and VectorMapNet (Camera) to demonstrate that LiDAR information can complement visual information to generate more robust map predictions.
In the first case, the camera view is constrained by the nearby vehicles, so it can not provide helpful surrounding information. LiDAR sensor bypasses the nearby vehicle and provides some cue for VectorMapNet to generate a better result than its camera-only counterpart (see Figure~\ref{fig:qualitative_Fusion1}). 
For the second case (see Figure~\ref{fig:qualitative_Fusion2}), the model cannot detect the nearby parking gate because it locates in the blind zone of cameras. In contrast, the LiDAR provides depth information and helps the VectorMapNet(Fusion) detect the missing lane boundary. 
LiDAR points can prevent the model from falsely detecting map elements in bad weather conditions as well. As shown in Figure~\ref{fig:qualitative_Fusion3}, some puddles are near the intersection. With the light reflection, these puddles visually look like a lane boundary. However, the LiDAR data shows that there does not have any bump in there. Unlike the camera-only model, this depth information from LiDAR helps our fusion model not generate a non existed lane boundary. 

\section{Implementation details}
\label{sec:implementation}
\subsection{Overall Architectures.} BEV feature extractor outputs a feature map with a size of $(200,100,128)$. It uses ResNet50~\citep{he2016deep} for shared CNN backbone. We use a single layer PointNet~\citep{qi2017pointnet} whose outputs have $64$ dimensions as the LiDAR backbone to aggregate LiDAR points into a pillar. 
We set the number of element queries $N_{\mathrm{max}}$ in map element detector as $100$. 
The transformer decoders we used in map element detector and polyline generator both have $6$ decoder layers, and their hidden embeddings' size is $256$. 
For the output space of polyline generator, we divide the map space (see \S~\ref{sec:gen}) evenly into $200\times100$ rectangular grids, and each grid has a size of $0.3m\times0.3m$.

\subsection{Training settings.}
We train all our models on 8 GTX3090 GPUs for 110 epochs with a total batch size of 32.
We use AdamW~\citep{loshchilov2018fixing} optimizer with a gradient clipping norm of $5.0$. For the learning rate schedule, we use a step schedule that multiplies a learning rate by 0.1 at epoch 100 and has a linear warm-up period at the first 5000 steps. The dropout rate for all modules is 0.2, following the transformer's settings~\citep{vaswani2017attention}. 
Data augmentation is only deployed during polyline generator's training; specifically, two I.I.D. Gaussian noises are added to each input vertex's $x$ and $y$ coordinates with a probability of $0.3$.

\subsection{Model Details}
\label{model_details}
\noindent\textbf{Camera Branch of Map Feature Extractor.}
For image data $\mathcal{I}$, we use a shared CNN backbone to obtain each camera's image features in the camera space, then use the Inverse Perspective Mapping~(IPM)~\citep{mallot1991inverse} technique to transform these features into BEV space. Since the depth information is missing in camera images, we follow one common approach that assumes the ground is mostly planar and transforms the images to BEV via homography. 
Without knowing the exact height of the ground plane, this homography is not an accurate transformation. 
To alleviate this issue, we transform the image features into four BEV planes with different heights ( we use ($-1m,0m,1m,2m$) in practice). The camera BEV features $\bm{\mathcal{F}}_{\mathrm{BEV}}^\mathcal{I}\in\mathbb{R}^{W \times H \times C_1}$ are the concatenation of these feature maps.

\subsection{Loss}
\label{subsec:loss}
\noindent\textbf{Loss settings.} The loss function of map element detector is a linear combination of three parts: a negative log-likelihood for element keypoint classification, a smooth L1 loss, and an IoU loss for keypoints regression. The coefficients of these loss components are $2,0.1,1$. The matching cost of map element detector is the same as the loss combination.
The loss function of polyline generator is a negative log-likelihood. 
We train VectorMapNet by simply summing up these losses.

\noindent\textbf{map element detector loss.}
To get the loss, we first establish a correspondence between the ground-truth ($\mathcal{A}$, $\bm{\mathcal{L}}$) and the prediction ($\hat{\mathcal{A}}$, $\hat{\bm{\mathcal{L}}}$). Assuming the number of ground-truth map element keypoints $N$ is smaller than the number of predictions $N_{max}$, and we pad the set of ground-truth ($\mathcal{A}$, $\bm{\mathcal{L}}$) with $\emptyset$s (no object) up to $N_{max}$. The correspondence $\sigma$ is a permutation of $N_{max}$ elements $\sigma\in\mathcal{P}$ with the lowest cost: 
 $\sigma^{\ast} = \underset{\sigma\in \mathcal{P}}{argmin} \sum_{j=1}^{N_{max}} -\mathds{1}_{(l_j \neq \emptyset)} \hat{p}_{\sigma(j)} (l_j) + -\mathds{1}_{(l_j \neq \emptyset)}\bm{\mathcal{L}}_{keypoint}(a_j, \hat{a}_{\sigma(j)})$, where $\hat{p}_{\sigma(j)}(l_j)$ is the probability of class label $l_j$ for the prediction with index $\sigma(j)$, and the loss of keypoints parameters $\bm{\mathcal{L}}_{keypoint}$ is an addition of a smooth L1 loss and an IoU loss. With these notations we define the loss of detector as: 
 $$\bm{\mathcal{L}}_{det} = \sum_{j=1}^{N_{max}} -\log \hat{p}_{\sigma^{\ast}(j)}(l_j) + \mathds{1}_{(l_j \neq \emptyset)} \bm{\mathcal{L}}_{keypoint}(a_j,\hat{a}_{\sigma^{\ast}(j)}),$$ where $\sigma^{\ast}$ is the optimal assignment computed by Hungarian algorithm~\citep{kuhn1955hungarian}.

 \subsection{Baseline model}
 \label{sec: baseline model settings}
 \noindent\textbf{HDMapNet}
 For all experiments in our paper, we employed the official HDMapNet model from the provided codebase and directly take its vectorized results. As the Argoverse dataset was not included in the original HDMapNet paper, we adapted the NuScenes data processing steps from their codebase to create an Argoverse2 dataloader for our experiments. 
 
\noindent\textbf{STSU}
For STSU, It uses a transformer module to detect the moving objects and centerline segments. It uses an association head to piece the segments together as the road graph. In order to adapt STSU to our task, we use a two-layer MLP to predict lane segments and only keep its object branch and polyline branch.

\section{More Ablation Studies}
\label{sec:more_ablations}
\subsection{Curve sampling strategies}
\begin{table}[h] \centering
    \ra{1.3}
    \caption{ 
        Ablation study of curves sampling strategies.
    }
    \resizebox{\textwidth}{!}{
    \begin{tabular}{c|llllcllll}
    \toprule
    \multicolumn{1}{c}{} & \multicolumn{4}{c}{Fr\'echet Distance} && \multicolumn{4}{c}{Chamfer Distance} \\ \cmidrule(lr){2-5} \cmidrule(lr){7-10}

    Vertex Sampling Method & AP$_{ped}$ & AP$_{divider}$ & AP$_{boundary}$ & mAP && AP$_{ped}$ & AP$_{divider}$ & AP$_{boundary}$ & mAP \\ \hline
    curvature-based &47.0 &47.4 &56.9 &50.4 &&27.6 &34.4 &35.4 &32.5 \\
    fixed interval  &26.0 &23.6 &37.1 &28.9 &&14.6 &17.6 &18.7 &17.0 \\
    \bottomrule
    \end{tabular}
    }
    \label{tab:polyline_representation}
\end{table}
We use two approaches to sample polylines. The first is based on the original nuScenes setting~\citep{caesar2020nuscenes}, which samples vertices at the position where the curvature changes are beyond a certain threshold. The second is to sample the vertices at fixed intervals ($1m$). We compare our methods under these two sampling strategies and the results are shown in Table~\ref{tab:polyline_representation}.
The curvature-based sampling outperforms its fixed-sampling counterpart by a large margin and achieves a leading 21.5 Fr\'echet mAP and 15.5 Chamfer mAP. We hypothesize that the fixed-sampling method involves a large set of redundant vertices that have negligible contributions to the geometry, thus under-weighs the essential vertices (\eg the vertices at the corner of a polyline) in the learning process.

\subsection{Vertex modeling methods.} 
\label{subsec:vertexmodeling}
\begin{table}[h] \centering
    \ra{1.3}
    \caption{ 
        Ablation study of vertex modeling methods. 
    }
    \resizebox{\textwidth}{!}{
    \begin{tabular}{c|llllcllll}
    \toprule
    \multicolumn{1}{c}{} & \multicolumn{4}{c}{Fr\'echet Distance} && \multicolumn{4}{c}{Chamfer Distance} \\ \cmidrule(lr){2-5} \cmidrule(lr){7-10}

    Modeling Method & AP$_{ped}$ & AP$_{divider}$ & AP$_{boundary}$ & mAP && AP$_{ped}$ & AP$_{divider}$ & AP$_{boundary}$ & mAP \\ \hline
    discrete &47.0 &47.4 &56.9 &50.4 &&27.6 &34.4 &35.4 &32.5 \\
    continuous &38.0 &41.6 &46.1 &41.9 &&26.5  &28.1 &30.1 &26.5 \\
    \bottomrule
    \end{tabular}
    }
    \label{tab:polyline_model}
\end{table}
We investigate both discrete and continuous ways to model polyline vertices. The discrete version of polyline generator is described in \S~\ref{sec:gen}. With the same model structure, we follow SketchRNN~\citep{ha2017neural} and use mixture of Gaussian distributions to model the vertices of polylines as continuous variables. The comparison is shown in Table~\ref{tab:polyline_model}.
We find that using discrete embeddings vertex coordinates results in a considerable gain in performance, with Chamfer mAP increasing from 18.2 to 32.5 and the Fr\'echet mAP increasing from 26.8 to 50.4. 
These improvements suggest that the non-local characteristic of categorical distribution helps our model to capture complex vertex coordinate distributions.

\subsection{Extrinsic Robustness}
Thanks for this suggestion. To probe the robustness of VectorMapNet, we follow lift-splat-shoot\cite{philion2020lift}, which tests the model under the noise that occurs in self-driving, such as camera extrinsic being biased.
The table~\ref{tab:robustness} shows that training the model with noisy extrinsic can lead to better test-time performance. And our model maintains its good performance for high amounts of extrinsic noise. The results show our model's robustness against extrinsic noise.

\begin{table}[!ht]
    \centering
    \caption{VectorMapNet performance under different extrinsic noise.}
    \scalebox{1.}{
    \begin{tabular}{@{}c|llll@{}}
    \hline
        mAP & \multicolumn{4}{c}{Test time extrinsic noise} \\ \hline
        Train time extrinsic noise & 0 & 0.1 & 0.3 & 0.6 \\ 
        0 & 42.2 & 42.6 & 42.4 & 42.5 \\ 
        0.1 & \textbf{43.8} & 43.6 & 43.5 & 43.6 \\ 
        0.3 & 43.0 & 43.0 & 43.1 & 43.1 \\ 
        0.6 & 42.6 & 42.72 & 42.8 & 42.9 \\ \hline
    \end{tabular}
    }
    \label{tab:robustness}
\end{table}

\section{Additional Discussions}
\label{sec:Discussions}

\noindent\textbf{The potential negative societal impact.} While there are legitimate concerns regarding privacy issues in autonomous driving systems that use generated maps, we'd like to reassure that our method is designed with such concerns in mind. Our proposed VectorMapNet model relies solely on onboard sensor observations and doesn't track any global locations or individual movements. Consequently, it poses no risk of leaking personal information, such as patterns in individuals' movements, thus upholding the highest standards of privacy.

\noindent\textbf{Confidence indicator.} Learning-based models can certainly provide confidence indicators to describe prediction uncertainty. Our model can generate two types of confidence scores: (1) The DETR-like Map Element Detector generates a confidence score for each detected map element. It is an instance-level score. (2) The auto-regressive Polyline Generator generates a score for each point on a polyline. It is a point-level score. Both of them can indicate the confidence or likelihood of the model's prediction. However, how to better use this uncertainty for downstream tasks still remains an open question. We leave it for future research.

\end{document}